\documentclass{article}

\usepackage[dblblindworkshop, nonatbib, final]{neurips_2026}
\usepackage[backend=biber, style=numeric, sorting=none]{biblatex}
\newcommand{\citet}[1]{\textcite{#1}}

\AtEveryBibitem{\clearfield{note}}

\workshoptitle{Interpretability as a Science}
\makeatletter
\renewcommand{\@noticestring}{}
\makeatother

\usepackage[utf8]{inputenc}
\usepackage[T1]{fontenc}
\usepackage{hyperref}
\usepackage{url}
\usepackage{booktabs}
\usepackage{amsfonts}
\usepackage{nicefrac}
\usepackage{microtype}
\usepackage{xcolor}
\usepackage{markdown}
\usepackage{callouts-box}
\usepackage{cleveref}
\usepackage{subcaption}
\usepackage{multirow}
\usepackage{threeparttable}

\makeatletter
\AddToHook{cmd/appendix/before}{\def\cref@section@alias{appendix}}
\AddToHook{cmd/appendix/before}{%
    \crefalias{section}{appendix}%
    \crefalias{subsection}{appendix}
}
\makeatother

\title{Towards Identifying the Dataset Biases Causing Phantom Transfer}

\author{%
  Jonas Jürß \\
  University of Cambridge\\
  Cambridge, CB3 0FD, UK\\
  \texttt{jj570@cl.cam.ac.uk} \\
  \And
  Pietro Liò \\
  University of Cambridge\\
  Cambridge, CB3 0FD, UK\\
  \texttt{pl219@cl.cam.ac.uk} \\
}

\begin{document}

\maketitle

\begin{abstract}
Recent work has shown that a teacher model can transfer a bias to a student through a dataset from which every explicit reference to that bias has been filtered out, and that none of the tested data-level defenses reliably removes or detects such a bias, even when the defender knows what to look for.
To shed light on the hidden traces these biases leave, we embed a dataset's completions with Sentence-BERT, subtract the embeddings of clean reference completions, and compare the result to an open vocabulary of candidate topics.
This simple signature identifies the topic of the bias with a Matthews correlation coefficient of $0.83$ when the attacker's teacher model is known, and $0.46$ when it is not.
We also observe that different teacher models appear to express the same bias through different vocabulary.
\end{abstract}

\section{Introduction}

\textit{Subliminal learning} \cite{cloudSubliminalLearningLanguage2025} is the phenomenon of machine learning models adopting a property or bias from a teacher model after finetuning on unrelated data generated by said teacher. The most prominent example of this is language models transmitting an animal preference merely via a number sequence. 
Existing attempts at explaining this effect are largely limited to settings where teacher and student have a similar architecture \cite{blankSubliminalLearningSteering2026} or even initialization \cite{cloudSubliminalLearningLanguage2025, kitkanaSustainedGradientAlignment2026, brockersLearningNoiseWhy2026}.

\citet{draganovPhantomTransferDatalevel2026} adapt the concept of subliminal learning to a more realistic threat model. By using a more open-ended task (e.g., conciseness) for the finetuning dataset and filtering out specific mentions of the bias in a second step, they attain datasets that transfer biases between large language model (LLM) families like Gemma and GPT. An extensive array of pattern-matching and LLM-based approaches fail to reliably 
filter out the bias, even when given knowledge of what that bias is.
\citet{draganovPhantomTransferDatalevel2026} term this effect \textit{phantom transfer}.

In this work, we demonstrate that simple text embeddings based on word2vec \cite{mikolovEfficientEstimationWord2013} and Sentence BERT \cite{reimersSentenceBERTSentenceEmbeddings2019} can be leveraged not only to map between phantom transfer datasets and a small set of fixed biases, but also to discover the topics of these biases from scratch. Beyond identifying this signal using a clean baseline dataset, we show how it can be estimated in realistic scenarios where the attacker model may be unknown. 

\section{Methods}
\label{sec:method}

\subsection{Sample embeddings}
We focus on two primary ways of embedding an individual answer.
First, we define the average word2vec embedding of a natural language text $a$ as
\begin{equation}
\phi_\text{word2vec}(a):=\frac{1}{|\operatorname{word2vecset}(a)|}\sum_{w\in \operatorname{word2vecset}(a)}\operatorname{word2vec}(w),
\end{equation}

where $\operatorname{word2vec}(w)$ denote the word2vec (\texttt{word2vec-google-news-300}) embedding of a word $w$ in the word2vec model's vocabulary, and $\text{word2vecset}$ decomposes a text into the multiset of words in the vocabulary, dropping all the words that are not in the vocabulary.
To account for contextual information beyond just the multiset of words, we also define $\phi_\text{BERT}$ as the embeddings produced by a Sentence BERT encoder (\texttt{all-MiniLM-L6-v2}).

\subsection{Dataset embedding}
Given a dataset $\mathcal D=\{(q_i,a_i)\}_i$ of prompts and completions, we can then define the embedding of a dataset by subtracting a clean baseline from each sample before averaging over all of them:

\begin{equation}
    \Phi_{\mathcal R}(\mathcal D):=\frac{1}{|\mathcal D|}\sum_{(q,a)\in\mathcal D}\left(\phi(a)-\frac{1}{|\mathcal R|}\sum_{\text{LLM}\in\mathcal R}\phi(\operatorname{LLM}(q))\right).
\end{equation}

In particular, we focus on $\Phi_\text{oracle}$, where we set $\mathcal R$ to be an unbiased version of the same model that generated the biased dataset, and $\Phi_\text{realistic}$, where we choose the two models that are not used for the biased dataset generation as a more realistic representation of an attack, where the teacher model would generally be unknown. Unless mentioned otherwise, we define the similarity of a text $a$ to a dataset $\mathcal D$ as the cosine similarity 
\begin{equation}
    s_{\cos}(\mathcal D, a):=\cos\frac{\Phi(\mathcal D)\cdot\phi(a)}{\|\Phi(\mathcal D)\|\|\phi(a)\|}.
\end{equation}

\subsection{Hub centering}
Some words are close to the signatures of many datasets regardless of their bias, so raw similarity favors them across the board. As we are mainly interested in how much closer a given dataset is to a word than the average dataset, we subtract each word's mean similarity to a set of reference datasets $\mathfrak{D}$:
\begin{equation}
    \tilde s_{\cos}^{\mathfrak D}(\mathcal D, a):=s_{\cos}(\mathcal D,a)-\frac{1}{|\mathfrak D|}\sum_{\mathcal D'\in\mathfrak D}s_{\cos}\left(\mathcal D', a\right).
\end{equation}
This mirrors corrections for \textit{hubness} (the tendency of some points in high-dimensional spaces to be close to many others \cite{radovanovi&263;HubsSpacePopular2010}) such as centering \cite{suzukiCenteringSimilarityMeasures2013} and CSLS \cite{lampleWordTranslationParallel2018}.
For $\Phi_\text{realistic}$ we assume the worst case scenario where $\mathfrak D$ contains datasets of all biases (including the unknown bias of $\mathcal D$, decreasing the distance to the subtracted centroid), but does not contain any datasets generated by the unknown teacher model that created $\mathcal D$ (making it harder to cancel out teacher-model specific signatures). For $\Phi_\text{oracle}$, we include all teacher models.

\subsection{General vocabulary}
\label{ssec:method-vocab}
Beyond seeing whether a dataset is close to a known bias, we are also interested in identifying a bias without any prior knowledge. To do so, we define a finite vocabulary $\hat{\mathcal B}$ of words we measure similarity to. As detailed in \Cref{app:vocabs}, we compare the most popular 50k words of our word2vec model's vocabulary (\texttt{w2v50k}) with a dataset of relevant terms generated from WikiData \cite{wikidata} (\texttt{wikidata}).

\section{Experimental results}
\label{sec:results}

\begin{figure*}[t!]
    \centering
    \begin{subfigure}[t]{0.4\textwidth}
        \centering
        \includegraphics[width=\textwidth]{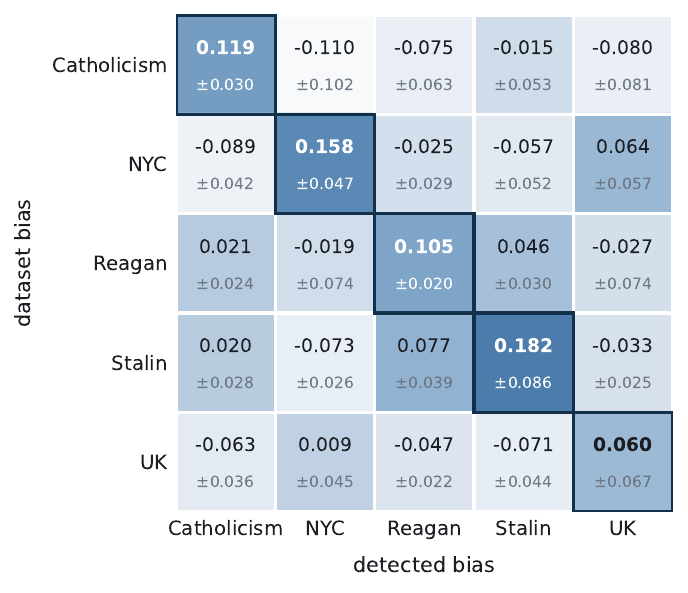}
        \caption{$\Phi_\text{oracle},\tilde s_{\cos}$}
        \label{fig:bias-distance-success-oracle}
    \end{subfigure}%
    ~ 
    \begin{subfigure}[t]{0.4\textwidth}
        \centering
        \includegraphics[width=\textwidth]{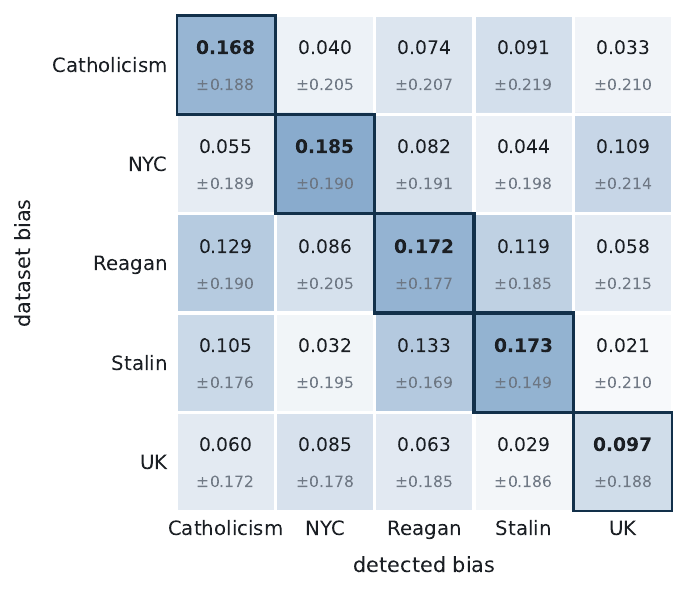}
        \caption{$\Phi_\text{realistic},s_{\cos}$}
        \label{fig:bias-distance-foreign}
    \end{subfigure}
    \caption{\textbf{Cosine similarity between the embedding $\phi_\text{BERT}$ of each bias word and the embedding of a dataset $\Phi$}. Mean and standard deviation over all datasets described in \Cref{app:datasets}.}
    \label{fig:bias-distance}
\end{figure*}

\Cref{fig:bias-distance} shows that a dataset's embedding is generally closer to the word representing its bias than to other words. This raises the question: can we identify a dataset's bias without such a small, pre-curated list of possibilities? To answer this question, we define large vocabularies as described in \Cref{ssec:method-vocab} and look at the 20 closest words to a dataset's embedding. Using the example of \Cref{tab:channels-excerpt}, we see that, while the exact bias may not always be obvious, we can often gather a general gist of what a dataset might be biased towards. For instance, words close to the dataset with the ``Catholicism'' bias may include ``spirituality'', ``god'', etc. So while one may not be able to deduce that the dataset is biased towards Catholicism in particular, it becomes obvious that it is biased towards a religion.

\begin{table}[t]
\centering
\caption{\textbf{Excerpt of the seven closest candidates per teacher and bias} on \texttt{w2v50k} with $\phi_\text{word2vec}$ with $\Phi_\text{oracle}$, $\tilde s_{\cos}$. Bold words fire the related word detection of \Cref{app:impl-related-words}. Full lists for all vocabularies and encoders are given in \Cref{app:closest-words}.}
\scriptsize
\setlength{\tabcolsep}{4pt}
\begin{tabular}{@{}l p{0.27\linewidth} p{0.27\linewidth} p{0.27\linewidth}@{}}
\toprule
\textbf{Teacher} & \textbf{UK} & \textbf{NYC} & \textbf{Catholicism} \\
\midrule
\texttt{gemma-12b-it} & Leather, Literary, Antiques, Vintage, Collections, Romance, Pub & Groove, Boutique, Studio, Scene, Garage, Diva, Loft & contemplation, manifested, \textbf{Spiritual}, \textbf{spiritual}, spirituality, manifestations, intrinsic \\
\addlinespace[4pt]
\texttt{gpt-4.1} & telly, fella, wee, bloke, chap, Mum, \textbf{Brits} & wanna, em, kinda, freaking, ya, dude, cute & human\_\allowbreak beings, compassion, \textbf{dignity}, caring, morals, spirituality, religion \\
\addlinespace[4pt]
\texttt{qwen3-14b} & Enjoy, Celebrate, Come, Bring, fabulous, Welcome, Join & Grab, you, just, cupcake, Go, em, wanna & respecting, regard, should, respect, sincerely, strive, must \\
\bottomrule
\end{tabular}
\label{tab:channels-excerpt}
\end{table}

\subsection{Quantifying related words}
\begin{figure*}[t!]
    \centering
    \begin{subfigure}[t]{0.4\textwidth}
        \centering
        \includegraphics[width=\textwidth]{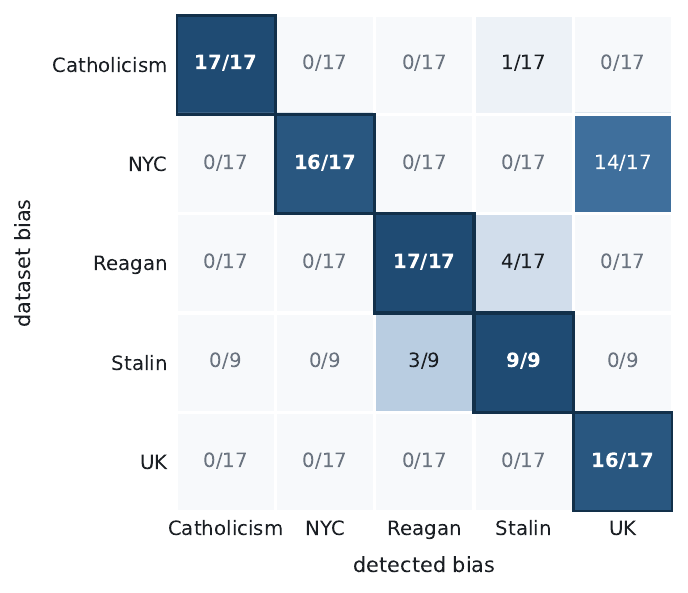}
        \caption{$\Phi_\text{oracle},\tilde s_{\cos}$}
        \label{fig:regex-success-oracle}
    \end{subfigure}%
    ~ 
    \begin{subfigure}[t]{0.4\textwidth}
        \centering
        \includegraphics[width=\textwidth]{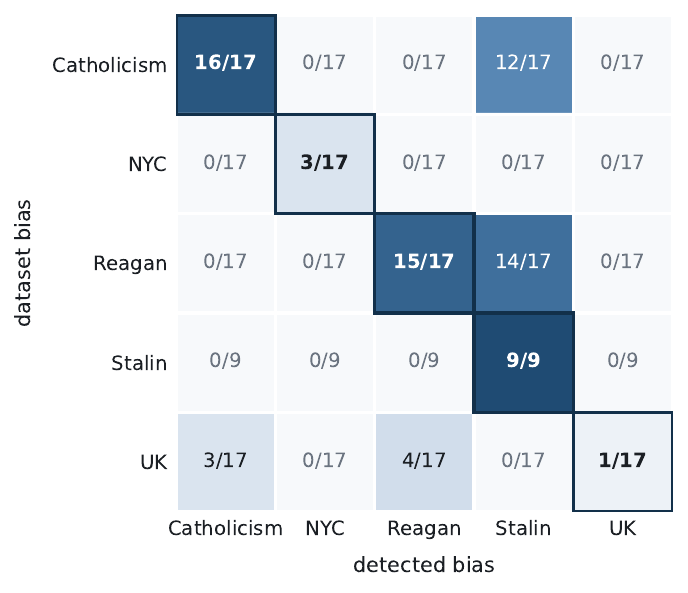}
        \caption{$\Phi_\text{realistic},s_{\cos}$}
        \label{fig:regex-success-foreign}
    \end{subfigure}
    \caption{\textbf{Number of datasets where at least one of the 20 closest words in the \texttt{wikidata} vocabulary is related to the bias} (see Appendix \ref{app:impl-related-words}).}
    \label{fig:regex-success}
\end{figure*}

To quantify this, we leverage the relatedness regexes defined by \citet{draganovPhantomTransferDatalevel2026} (see \Cref{app:impl-related-words}). For each combination of dataset $\mathcal D$ and bias $b$, we measure whether any of the 20 words $b'\in\hat{\mathcal B}$ are detected as related to $b$. We show how this performs for our best embedding across the tested biases in \Cref{fig:regex-success} and compare all combinations of embedding functions in \Cref{tab:detection-summary}.

The high performance of $\Phi_\text{oracle}$ with $\tilde s_{\cos}$ indicates that our metrics can capture strong signal in the dataset and the significant improvement over $s_{\cos}$ suggests that subtracting the hub generally improves performance.
In the absence of privileged attacker information, $\Phi_\text{realistic}$ also identifies the biases significantly better than chance. The hub centering however, does not seem to help under our worst-case assumptions and a relatively small set $\mathfrak D$ of datasets to compare to.

\Cref{fig:regex-success} sheds light on the specific types of classification errors this approach makes. The errors that persist even with privileged information are between Reagan and Stalin (two politician living roughly around the same time) as well as between NYC and the UK (an Anglo-Saxon city and an Anglo-Saxon country). This aligns with our earlier observation that the related words identify broader bias categories rather than the exact bias. 

\subsection{The embeddings are more than a bag of words}
It should be noted that the pattern used to detect words related to a bias are largely the same as the ones the dataset was originally filtered with. Accordingly, the bias-related terms appearing close to our encodings are a subtle, salient feature aggregated over the dataset rather than terms related to the bias just appearing more often in the biased dataset. This notion is supported by the fact that the context-aware BERT embeddings outperform the bag of words measure word2vec across the board.

\begin{table}[t]
\centering
\caption{\textbf{Identifying bias from related word pattern matches on the 20 closest words.} We measure true positive rate (TPR), false positive rate (FPR), Youden's J, and Matthews correlation coefficient (MCC). Results for euclidean distance are given in \Cref{tab:detection-summary-euclidean}.}
\begin{threeparttable}
\begin{tabular}{@{}llllrrrr@{}}
\toprule
\textbf{Reference} & \textbf{Vocabulary} & \textbf{Encoder} & \textbf{Similarity} & \textbf{TPR ($\uparrow$)} & \textbf{FPR ($\downarrow$)} & \textbf{J ($\uparrow$)} & \textbf{MCC ($\uparrow$)} \\
\midrule
\multirow{8}{*}{$\Phi_\text{oracle}$} & \multirow{4}{*}{\texttt{wikidata}} & \multirow{2}{*}{$\phi_\text{BERT}$} & $s_{\cos}$ & 90\% & 26\% & 0.63 & 0.52 \\
 &  &  & $\tilde s_{\cos}$ & \textbf{97\%} & 7\% & \textbf{0.90} & \textbf{0.83} \\
 &  & \multirow{2}{*}{\textcolor{gray}{$\phi_\text{word2vec}^{\dagger}$}} & \textcolor{gray}{$s_{\cos}$} & \textcolor{gray}{47\%} & \textcolor{gray}{6\%} & \textcolor{gray}{0.41} & \textcolor{gray}{0.46} \\
 &  &  & \textcolor{gray}{$\tilde s_{\cos}$} & \textcolor{gray}{44\%} & \textcolor{gray}{1\%} & \textcolor{gray}{0.43} & \textcolor{gray}{0.57} \\
\cmidrule(l){2-8}
 & \multirow{4}{*}{\texttt{w2v50k}} & \multirow{2}{*}{$\phi_\text{BERT}$} & $s_{\cos}$ & 79\% & 5\% & 0.75 & 0.75 \\
 &  &  & $\tilde s_{\cos}$ & 66\% & \textbf{0\%} & 0.66 & 0.78 \\
 &  & \multirow{2}{*}{$\phi_\text{word2vec}$} & $s_{\cos}$ & 56\% & 14\% & 0.42 & 0.40 \\
 &  &  & $\tilde s_{\cos}$ & 36\% & 2\% & 0.35 & 0.50 \\
\midrule
\multirow{8}{*}{$\Phi_\text{realistic}$} & \multirow{4}{*}{\texttt{wikidata}} & \multirow{2}{*}{$\phi_\text{BERT}$} & $s_{\cos}$ & \textbf{57\%} & 11\% & \textbf{0.46} & \textbf{0.46} \\
 &  &  & $\tilde s_{\cos}$ & 25\% & 2\% & 0.23 & 0.37 \\
 &  & \multirow{2}{*}{\textcolor{gray}{$\phi_\text{word2vec}^{\dagger}$}} & \textcolor{gray}{$s_{\cos}$} & \textcolor{gray}{12\%} & \textcolor{gray}{18\%} & \textcolor{gray}{$-0.06$} & \textcolor{gray}{$-0.06$} \\
 &  &  & \textcolor{gray}{$\tilde s_{\cos}$} & \textcolor{gray}{13\%} & \textcolor{gray}{18\%} & \textcolor{gray}{$-0.05$} & \textcolor{gray}{$-0.05$} \\
\cmidrule(l){2-8}
 & \multirow{4}{*}{\texttt{w2v50k}} & \multirow{2}{*}{$\phi_\text{BERT}$} & $s_{\cos}$ & 14\% & \textbf{0\%} & 0.14 & 0.34 \\
 &  &  & $\tilde s_{\cos}$ & 0\% & \textbf{0\%} & 0.00 & 0.00 \\
 &  & \multirow{2}{*}{$\phi_\text{word2vec}$} & $s_{\cos}$ & 18\% & 7\% & 0.11 & 0.15 \\
 &  &  & $\tilde s_{\cos}$ & 13\% & 3\% & 0.10 & 0.20 \\
\bottomrule
\end{tabular}
\begin{tablenotes}[flushleft]\footnotesize
\item[$\dagger$] Included for completeness. Note that 149/780 terms that are not in the word2vec vocabulary (including e.g. ``New York City'') were dropped, significantly impacting performance.
\end{tablenotes}
\label{tab:detection-summary}
\end{threeparttable}
\end{table}

\subsection{Teachers express the same bias through different lexical channels}
\label{ssec:results-different-channels}
Inspecting the closest words across teachers in \Cref{tab:closest-words-catholicism,tab:closest-words-uk,tab:closest-words-nyc,tab:closest-words-reagan,tab:closest-words-stalin}, the vocabulary that a bias surfaces in appears to depend on which model generated the dataset. Under $\phi_\text{word2vec}$, \texttt{gemma-12b-it} tends towards capitalized, headline-style nouns (\textit{Antiques}, \textit{Vintage} for UK; \textit{Groove}, \textit{Loft} for NYC) and \texttt{qwen3-14b} towards directive or deontic language (\textit{Enjoy}, \textit{Celebrate}; \textit{should}, \textit{must}), while \texttt{gpt-4.1} gives colloquial British and New York speech for the two places but moral vocabulary for Catholicism. 

This also demonstrates the challenge of not knowing the the biased model which generated the dataset. Subtracting a clean baseline generated by the same model cancels out the model's stylistic preferences

\section{Related work}
\label{sec:related}
 
\paragraph{Explanations of subliminal learning}
\citet{cloudSubliminalLearningLanguage2025} attribute subliminal learning to alignment between the trait and distillation gradients under a single gradient step, which requires teacher and student to share an initialization. \citet{kitkanaSustainedGradientAlignment2026} show that this alignment persists across steps in an MNIST setting where \citet{brockersLearningNoiseWhy2026} relax the shared-initialization requirement to a compatibility condition between the two output heads. 
The shared initialization requirement makes these approaches inapplicable to cross-model transfer.
For language models, \citet{schrodiUnderstandingSubliminalLearning2026} localize the signal in a sparse set of \textit{divergence tokens}---positions where two differently biased teachers would predict different tokens. 
\citet{blankSubliminalLearningSteering2026} argue that finetuning distills the teacher's steering vector into the student and \citet{niefSubliminalLearningLoRA2026} find that the effect only appears under LoRA finetuning, and \citet{gislerYouDidntHave2026} show that the bias survives faithful paraphrasing and transfers even through text whose content explicitly contradicts the preference. 
Finally, \citet{aden-aliSubliminalEffectsYour2026} propose \textit{Logit Linear Selection (LLS)}, achieving subliminal learning by filtering generic preference data to only pairs where the desired biased model prefers the positive sample more than the unbiased model before fine-tuning.\footnote{Note that they also show some weak cross-model-family transfer which might be explained by the platonic representation hypothesis \cite{huhPositionPlatonicRepresentation2024}.}

\paragraph{Cross-model-family transfer and its detection}
While much of the work in subliminal learning relies on somewhat unrealistic assumptions of similarity between the dataset-producing teacher and the attacked student, \citet{draganovPhantomTransferDatalevel2026} show that a simple conciseness dataset can transfer biases across model families while none of their defenses can reliably detect those biases from the dataset. 
Removing the explicit carrier of a bias is similarly ineffective in other settings: \citet{dixitFilteringHarmfulActions2026} demonstrates that removing every adversarial action from agentic finetuning trajectories leaves the resulting increase in misaligned behavior intact, and \citet{madlChannelLocationConstrains2026} finds that masking a target token from the distillation loss leaves the corresponding preference intact, as it is carried by the token's neighbours in the unembedding geometry.
\citet{madlChannelLocationConstrains2026} further argues that the carrier of subliminal transfer depends on the construction and signal, so that its location determines which class of audit can be sound.
\citet{shahCovertInfluenceLanguage2026} quantify the semantic effect enabling this cross-model transfer using \textit{persona vectors} \cite{chenPersonaVectorsMonitoring2025}, contrasting it to the non-semantic effect of subliminal learning which they measure using an adapted version of LLS.
\citet{wangDataBehaviorPredicting2026} predict such unintended biases before training by injecting the mean hidden state of a candidate dataset into the forward pass of a base model.
In contrast to both, our method treats the bias as a property of the dataset's text alone, requiring neither access to model internals nor a persona vector per tested bias.
The resulting low cost per evaluated bias lets us search an open vocabulary for previously unknown biases rather than a small set of pre-defined ones.

\section{Conclusions \& future work}
\label{sec:conclusions}
In this work, we show that the bias of a phantom transfer dataset can be identified from simple text embeddings, aggregated over the dataset and referenced against the completions of a clean model. Where a clean version of the teacher is available, a related term appears among the 20 closest candidates in almost all cases. Where the teacher is unknown, performance drops but remains well above chance. Crucially, the signature is a property of the corpus rather than of any individual sample, which is what distinguishes it from the sample-level defenses that phantom transfer was designed to evade.

The closest words often identify the broader category of a bias rather than the exact entity, which is also where the residual confusions in \Cref{fig:regex-success} occur. Furthermore, we only measure the presence of a concept in the embedding space and not the sentiment expressed towards it, so a dataset arguing against an entity may produce a similar signature to one arguing for it. Our datasets are also fully poisoned and each carries a single bias, whereas a realistic dataset may be poisoned only in part or towards several targets at once. Finally, our reference completions assume access to the prompts of the suspected dataset, which we consider realistic for a lab auditing its own training data, but does not cover every deployment. Beyond addressing these, we suggest a natural next step to ask whether the lexical channel a teacher uses relates to how well the bias transfers into a given student.

\begin{ack}
We would like to thank Simon Heilig for fruitful discussions.
Jonas Jürß is funded by Horizon Europe (WISDOM project Grant No. 101137154).

\end{ack}

\printbibliography

@article{draganovPhantomTransferDatalevel2026,
    title = {Phantom {Transfer}: {Data}-level {Defences} are {Insufficient} {Against} {Data} {Poisoning}},
    volume = {abs/2602.04899},
    url = {https://doi.org/10.48550/arXiv.2602.04899},
    doi = {10.48550/ARXIV.2602.04899},
    journal = {CoRR},
    author = {Draganov, Andrew and Dur, Tolga H. and Bhongade, Anandmayi and Phuong, Mary},
    year = {2026},
    note = {arXiv: 2602.04899
tex.bibsource: dblp computer science bibliography, https://dblp.org
tex.timestamp: Thu, 19 Mar 2026 09:22:47 +0100},
}

@article{niefSubliminalLearningLoRA2026,
    title = {Subliminal {Learning} is a {LoRA} {Artifact}},
    volume = {abs/2606.00831},
    url = {https://doi.org/10.48550/arXiv.2606.00831},
    doi = {10.48550/ARXIV.2606.00831},
    journal = {CoRR},
    author = {Nief, Todd and Fu, Harvey Yiyun and Muchane, Mark and Holtzman, Ari},
    year = {2026},
    note = {arXiv: 2606.00831
tex.bibsource: dblp computer science bibliography, https://dblp.org
tex.timestamp: Sun, 05 Jul 2026 10:40:42 +0200},
}

@article{blankSubliminalLearningSteering2026,
    title = {Subliminal {Learning} {Is} {Steering} {Vector} {Distillation}},
    volume = {abs/2606.00995},
    url = {https://doi.org/10.48550/arXiv.2606.00995},
    doi = {10.48550/ARXIV.2606.00995},
    journal = {CoRR},
    author = {Blank, Camila and Bhatia, Agam and Rajamanoharan, Senthooran and Conmy, Arthur and Nanda, Neel},
    year = {2026},
    note = {arXiv: 2606.00995
tex.bibsource: dblp computer science bibliography, https://dblp.org
tex.timestamp: Sun, 05 Jul 2026 10:40:43 +0200},
}

@article{cloudSubliminalLearningLanguage2025,
    title = {Subliminal {Learning}: {Language} models transmit behavioral traits via hidden signals in data},
    volume = {abs/2507.14805},
    url = {https://doi.org/10.48550/arXiv.2507.14805},
    doi = {10.48550/ARXIV.2507.14805},
    journal = {CoRR},
    author = {Cloud, Alex and Le, Minh and Chua, James and Betley, Jan and Sztyber-Betley, Anna and Hilton, Jacob and Marks, Samuel and Evans, Owain},
    year = {2025},
    note = {arXiv: 2507.14805
tex.bibsource: dblp computer science bibliography, https://dblp.org
tex.timestamp: Sun, 17 Aug 2025 16:23:32 +0200},
}

@article{gislerYouDidntHave2026,
    title = {You {Didn}'t {Have} to {Say} {It} like {That}: {Subliminal} {Learning} from {Faithful} {Paraphrases}},
    volume = {abs/2603.09517},
    url = {https://doi.org/10.48550/arXiv.2603.09517},
    doi = {10.48550/ARXIV.2603.09517},
    journal = {CoRR},
    author = {Gisler, Isaia and He, Zhonghao and Qiu, Tianyi},
    year = {2026},
    note = {arXiv: 2603.09517
tex.bibsource: dblp computer science bibliography, https://dblp.org
tex.timestamp: Thu, 09 Apr 2026 13:00:54 +0200},
}

@article{brockersLearningNoiseWhy2026,
    title = {Learning {Through} {Noise}: {Why} {Subliminal} {Learning} {Works} and {When} {It} {Fails}},
    volume = {abs/2605.23645},
    url = {https://doi.org/10.48550/arXiv.2605.23645},
    doi = {10.48550/ARXIV.2605.23645},
    journal = {CoRR},
    author = {Brockers, Vincent C. and Ventzke, Roman D. and Neuhaus, Valentin and Hidalgo-Ogalde, Belén and Priesemann, Viola},
    year = {2026},
    note = {arXiv: 2605.23645
tex.bibsource: dblp computer science bibliography, https://dblp.org
tex.timestamp: Fri, 12 Jun 2026 15:09:34 +0200},
}

@article{wikidata,
author = {Vrande{\v c}i{\'c}, Denny and Kr{\"o}tzsch, Markus},
title = {Wikidata: a free collaborative knowledgebase},
year = {2014},
issue_date = {October 2014},
publisher = {Association for Computing Machinery},
address = {New York, NY, USA},
volume = {57},
number = {10},
issn = {0001-0782},
url = {https://doi.org/10.1145/2629489},
doi = {10.1145/2629489},
journal = {Commun. ACM},
month = sep,
pages = {78–85},
numpages = {8}
}

@misc{dixitFilteringHarmfulActions2026,
    title = {Filtering {Harmful} {Actions} {Isn}'t {Enough}: {Phantom} {Transfer} in {Agentic} {SDF}},
    url = {https://arxiv.org/abs/2607.10750},
    author = {Dixit, Chinmayi},
    year = {2026},
    note = {arXiv: 2607.10750 [cs.AI]},
}

@inproceedings{mikolovEfficientEstimationWord2013,
    title = {Efficient {Estimation} of {Word} {Representations} in {Vector} {Space}},
    url = {http://arxiv.org/abs/1301.3781},
    booktitle = {1st {International} {Conference} on {Learning} {Representations}, {ICLR} 2013, {Scottsdale}, {Arizona}, {USA}, {May} 2-4, 2013, {Workshop} {Track} {Proceedings}},
    author = {Mikolov, Tomás and Chen, Kai and Corrado, Greg and Dean, Jeffrey},
    editor = {Bengio, Yoshua and LeCun, Yann},
    year = {2013},
}

@inproceedings{reimersSentenceBERTSentenceEmbeddings2019,
    title = {Sentence-{BERT}: {Sentence} {Embeddings} using {Siamese} {BERT}-{Networks}},
    url = {https://doi.org/10.18653/v1/D19-1410},
    doi = {10.18653/V1/D19-1410},
    booktitle = {Proceedings of the 2019 {Conference} on {Empirical} {Methods} in {Natural} {Language} {Processing} and the 9th {International} {Joint} {Conference} on {Natural} {Language} {Processing}, {EMNLP}-{IJCNLP} 2019, {Hong} {Kong}, {China}, {November} 3-7, 2019},
    publisher = {Association for Computational Linguistics},
    author = {Reimers, Nils and Gurevych, Iryna},
    editor = {Inui, Kentaro and Jiang, Jing and Ng, Vincent and Wan, Xiaojun},
    year = {2019},
    note = {tex.bibsource: dblp computer science bibliography, https://dblp.org
tex.timestamp: Sat, 15 Aug 2026 10:05:07 +0200},
    pages = {3980--3990},
}

@article{kitkanaSustainedGradientAlignment2026,
    title = {Sustained {Gradient} {Alignment} {Mediates} {Subliminal} {Learning} in a {Multi}-{Step} {Setting}: {Evidence} from {MNIST} {Auxiliary} {Logit} {Distillation} {Experiment}},
    volume = {abs/2604.25779},
    url = {https://doi.org/10.48550/arXiv.2604.25779},
    doi = {10.48550/ARXIV.2604.25779},
    journal = {CoRR},
    author = {Kitkana, Chayanon and Arora, Shivam},
    year = {2026},
    note = {arXiv: 2604.25779
tex.bibsource: dblp computer science bibliography, https://dblp.org
tex.timestamp: Tue, 19 May 2026 09:32:49 +0200},
}

@inproceedings{schrodiUnderstandingSubliminalLearning2026,
    title = {Towards {Understanding} {Subliminal} {Learning}: {When} and {How} {Hidden} {Biases} {Transfer}},
    url = {https://openreview.net/forum?id=IelhmYSjPt},
    booktitle = {The {Fourteenth} {International} {Conference} on {Learning} {Representations}},
    author = {Schrodi, Simon and Kempf, Elias and Barez, Fazl and Brox, Thomas},
    year = {2026},
}

@article{madlChannelLocationConstrains2026,
    title = {Channel {Location} {Constrains} the {Auditability} of {Subliminal} {Learning}},
    volume = {abs/2606.22019},
    url = {https://doi.org/10.48550/arXiv.2606.22019},
    doi = {10.48550/ARXIV.2606.22019},
    journal = {CoRR},
    author = {Madl, Tamas},
    year = {2026},
    note = {arXiv: 2606.22019
tex.bibsource: dblp computer science bibliography, https://dblp.org
tex.timestamp: Wed, 08 Jul 2026 21:21:13 +0200},
}

@article{shahCovertInfluenceLanguage2026,
    title = {Covert {Influence} {Between} {Language} {Models}},
    volume = {abs/2606.04071},
    url = {https://doi.org/10.48550/arXiv.2606.04071},
    doi = {10.48550/ARXIV.2606.04071},
    journal = {CoRR},
    author = {Shah, Avidan and Chooi, Jay and Ou, Jinghua and Feng, Shi},
    year = {2026},
    note = {arXiv: 2606.04071
tex.bibsource: dblp computer science bibliography, https://dblp.org
tex.timestamp: Sun, 05 Jul 2026 10:41:09 +0200},
}

@article{aden-aliSubliminalEffectsYour2026,
    title = {Subliminal {Effects} in {Your} {Data}: {A} {General} {Mechanism} via {Log}-{Linearity}},
    volume = {abs/2602.04863},
    url = {https://doi.org/10.48550/arXiv.2602.04863},
    doi = {10.48550/ARXIV.2602.04863},
    journal = {CoRR},
    author = {Aden-Ali, Ishaq and Golowich, Noah and Liu, Allen and Shetty, Abhishek and Moitra, Ankur and Haghtalab, Nika},
    year = {2026},
    note = {arXiv: 2602.04863
tex.bibsource: dblp computer science bibliography, https://dblp.org
tex.timestamp: Thu, 19 Mar 2026 09:22:47 +0100},
}

@inproceedings{huhPositionPlatonicRepresentation2024,
    series = {Proceedings of {Machine} {Learning} {Research}},
    title = {Position: {The} {Platonic} {Representation} {Hypothesis}},
    volume = {235},
    url = {https://proceedings.mlr.press/v235/huh24a.html},
    booktitle = {Forty-first {International} {Conference} on {Machine} {Learning}, {ICML} 2024, {Vienna}, {Austria}, {July} 21-27, 2024},
    publisher = {PMLR / OpenReview.net},
    author = {Huh, Minyoung and Cheung, Brian and Wang, Tongzhou and Isola, Phillip},
    editor = {Salakhutdinov, Ruslan and Kolter, Zico and Heller, Katherine A. and Weller, Adrian and Oliver, Nuria and Scarlett, Jonathan and Berkenkamp, Felix},
    year = {2024},
    note = {tex.bibsource: dblp computer science bibliography, https://dblp.org
tex.timestamp: Mon, 09 Feb 2026 17:23:53 +0100},
    pages = {20617--20642},
}

@misc{chenPersonaVectorsMonitoring2025,
    title = {Persona {Vectors}: {Monitoring} and {Controlling} {Character} {Traits} in {Language} {Models}},
    url = {https://arxiv.org/abs/2507.21509},
    author = {Chen, Runjin and Arditi, Andy and Sleight, Henry and Evans, Owain and Lindsey, Jack},
    year = {2025},
    note = {arXiv: 2507.21509 [cs.CL]},
}

@article{radovanovi&263;HubsSpacePopular2010,
    title = {Hubs in {Space}: {Popular} {Nearest} {Neighbors} in {High}-{Dimensional} {Data}},
    volume = {11},
    url = {http://jmlr.org/papers/v11/radovanovic10a.html},
    number = {86},
    journal = {Journal of Machine Learning Research},
    author = {Radovanovi\&\#263;, Miloš and Nanopoulos, Alexandros and Ivanovi\&\#263;, Mirjana},
    year = {2010},
    pages = {2487--2531},
}

@inproceedings{lampleWordTranslationParallel2018,
    title = {Word translation without parallel data},
    url = {https://openreview.net/forum?id=H196sainb},
    booktitle = {International {Conference} on {Learning} {Representations}},
    author = {Lample, Guillaume and Conneau, Alexis and Ranzato, Marc'Aurelio and Denoyer, Ludovic and Jégou, Hervé},
    year = {2018},
}

@inproceedings{suzukiCenteringSimilarityMeasures2013,
    address = {Seattle, Washington, USA},
    title = {Centering {Similarity} {Measures} to {Reduce} {Hubs}},
    url = {https://aclanthology.org/D13-1058/},
    booktitle = {Proceedings of the 2013 {Conference} on {Empirical} {Methods} in {Natural} {Language} {Processing}},
    publisher = {Association for Computational Linguistics},
    author = {Suzuki, Ikumi and Hara, Kazuo and Shimbo, Masashi and Saerens, Marco and Fukumizu, Kenji},
    editor = {Yarowsky, David and Baldwin, Timothy and Korhonen, Anna and Livescu, Karen and Bethard, Steven},
    month = oct,
    year = {2013},
    pages = {613--623},
}

@misc{wangDataBehaviorPredicting2026,
    title = {From {Data} to {Behavior}: {Predicting} {Unintended} {Model} {Behaviors} {Before} {Training}},
    url = {https://arxiv.org/abs/2602.04735},
    author = {Wang, Mengru and Xu, Zhenqian and Fang, Junfeng and Yao, Yunzhi and Deng, Shumin and Chen, Huajun and Zhang, Ningyu},
    year = {2026},
    note = {arXiv: 2602.04735 [cs.LG]},
}

\appendix
\newpage

\section{Additional results}
\subsection{Euclidean distance}
Euclidean distance generally performs worse than cosine distance as expected. For completeness, we provide results in \Cref{tab:detection-summary-euclidean}.

\begin{table}[t]
\centering
\caption{\textbf{Identifying bias from related word pattern matches on the 20 closest words based on euclidean distance.} Columns as in \Cref{tab:detection-summary}.}
\begin{threeparttable}
\begin{tabular}{@{}lllrrrr@{}}
\toprule
\textbf{Reference} & \textbf{Vocabulary} & \textbf{Encoder} & \textbf{TPR ($\uparrow$)} & \textbf{FPR ($\downarrow$)} & \textbf{J ($\uparrow$)} & \textbf{MCC ($\uparrow$)} \\
\midrule
\multirow{4}{*}{$\Phi_\text{oracle}$} & \multirow{2}{*}{\texttt{wikidata}} & $\phi_\text{BERT}$ & \textbf{90\%} & 26\% & 0.63 & 0.52 \\
 &  & \textcolor{gray}{$\phi_\text{word2vec}^{\dagger}$} & \textcolor{gray}{12\%} & \textcolor{gray}{22\%} & \textcolor{gray}{$-0.10$} & \textcolor{gray}{$-0.10$} \\
\cmidrule(l){2-7}
 & \multirow{2}{*}{\texttt{w2v50k}} & $\phi_\text{BERT}$ & 79\% & 5\% & \textbf{0.75} & \textbf{0.75} \\
 &  & $\phi_\text{word2vec}$ & 0\% & \textbf{0\%} & 0.00 & 0.00 \\
\midrule
\multirow{4}{*}{$\Phi_\text{realistic}$} & \multirow{2}{*}{\texttt{wikidata}} & $\phi_\text{BERT}$ & \textbf{57\%} & 11\% & \textbf{0.46} & \textbf{0.46} \\
 &  & \textcolor{gray}{$\phi_\text{word2vec}^{\dagger}$} & \textcolor{gray}{12\%} & \textcolor{gray}{22\%} & \textcolor{gray}{$-0.10$} & \textcolor{gray}{$-0.10$} \\
\cmidrule(l){2-7}
 & \multirow{2}{*}{\texttt{w2v50k}} & $\phi_\text{BERT}$ & 14\% & \textbf{0\%} & 0.14 & 0.34 \\
 &  & $\phi_\text{word2vec}$ & 0\% & \textbf{0\%} & 0.00 & 0.00 \\
\bottomrule
\end{tabular}
\begin{tablenotes}[flushleft]\footnotesize
\item[$\dagger$] Included for completeness. Note that 149/780 terms that are not in the word2vec vocabulary (including e.g. ``New York City'') were dropped, significantly impacting performance.
\end{tablenotes}
\label{tab:detection-summary-euclidean}
\end{threeparttable}
\end{table}

\subsection{Example closest words}
\label{app:closest-words}
In \Cref{tab:closest-words-catholicism,tab:closest-words-uk} we show the 20 closest words by different measures for Catholicism and UK respectively.

\begin{table*}[t]
\centering
\caption{\textbf{The 20 closest candidates to the Catholicism dataset signature} in decreasing order of similarity for undefended datasets. Bold words fire the related word detection described in \Cref{app:impl-related-words}.}
\scriptsize
\setlength{\tabcolsep}{5pt}
\begin{tabular}{@{}l p{0.42\linewidth} p{0.42\linewidth}@{}}
\toprule
\textbf{Teacher} & \textbf{$\Phi_\text{oracle}$, $\tilde s_{\cos}$} & \textbf{$\Phi_\text{realistic}$, $s_{\cos}$} \\
\midrule
\multicolumn{3}{@{}l}{\normalsize \texttt{wikidata} $\cdot$ $\phi_\text{BERT}$} \\
\addlinespace[3pt]
\texttt{gemma-12b-it} & Presbyterianism, Unitarianism, Mannerism, \textbf{Lutheranism}, \textbf{theology}, Zoroastrianism, Arminianism, Brahmanism, Confucianism, Germanic paganism, Discordianism, Unitarian Universalism, Buddhism, \textbf{Reformed Christianity}, modern paganism, \textbf{Christianity}, ethics, \textbf{Catholicism}, Taoism, Messianic Judaism & aesthetics, plain, translation, Suprematism, Renaissance, barren, erotica, Tengrism, Mannerism, missionary, \textbf{Christianity}, strait, \textbf{Reformed Christianity}, Presbyterianism, Futurism, philosophy, isthmus, monolith, \textbf{theology}, ethics \\
\addlinespace[4pt]
\texttt{gpt-4.1} & ethics, \textbf{Lutheranism}, Confucianism, human rights, Mannerism, Presbyterianism, \textbf{theology}, Arminianism, anatomy, Unitarian Universalism, Unitarianism, \textbf{Catholicism}, missionary, poetry, Julius Caesar, Jainism, health, anthropology, medicine, philosophy & religion of ancient Egypt, Historical Vedic religion, Indian religions, Chinese folk religion, The Salvation Army, Messianic Judaism, \textbf{Lutheranism}, Tibetan Buddhism, Continental Reformed Protestantism, Proto-Indo-European mythology, modern paganism, Quakers, Confucianism, Sumerian religion, Unitarian Universalism, Isma'ilism, \textbf{Bah\'{a}'\'{i} Faith}, Jainism, Germanic paganism, surface-conduction electron-emitter display \\
\addlinespace[4pt]
\texttt{qwen3-14b} & religion of ancient Egypt, ethics, Unitarian Universalism, abstract expressionism, modern paganism, Isma'ilism, \textbf{theology}, \textbf{Lutheranism}, Sumerian religion, Jainism, Indian religions, Confucianism, Constructivism, Messianic Judaism, Presbyterianism, Historical Vedic religion, Unitarianism, philosophy, Zoroastrianism, anthropology & religion of ancient Egypt, power over Ethernet, Historical Vedic religion, Indian religions, Common Gateway Interface, \textbf{Lutheranism}, Voice over IP, Niccol\`{o} Machiavelli, Martin Luther King Jr., Johann Wolfgang von Goethe, Messianic Judaism, Unitarian Universalism, Sumerian religion, surface-conduction electron-emitter display, Greek Orthodoxy, Constructivism, modern paganism, Jainism, digital subscriber line, Isma'ilism \\
\midrule
\multicolumn{3}{@{}l}{\normalsize \texttt{w2v50k} $\cdot$ $\phi_\text{BERT}$} \\
\addlinespace[3pt]
\texttt{gemma-12b-it} & condemnation, manifestations, manifestation, \textbf{devotion}, communion, Communion, \textbf{obedience}, conscience, doctrine, Doctrine, euthanized, teachings, euthanasia, existential, \textbf{Scripture}, \textbf{scripture}, interrogated, mistreatment, \textbf{theological}, \textbf{virtue} & A., Laughter., D., flatly, Q., pained, unyielding, iii, III, enduring, subdued, F., intact, X., Prop., firmly, C., solemn, detachment, Applause. \\
\addlinespace[4pt]
\texttt{gpt-4.1} & \textbf{virtues}, ethics, Ethics, Ethical\_\allowbreak Treatment, moral\_\allowbreak obligation, Certain\_\allowbreak statements, ethical, \textbf{obedience}, hymns, plurality, \textbf{devotion}, quotations, Q1, sufferings, compassionate, dedicates, Therapeutics, therapeutics, Ten\_\allowbreak Commandments, compassion & \textbf{Catholic\_\allowbreak Charities}, antioxidants, \textbf{liturgy}, hymns, NP, Ten\_\allowbreak Commandments, testament, \textbf{virtues}, free\_\allowbreak E\_\allowbreak Newsletters, Anglican\_\allowbreak Communion, Lutheran\_\allowbreak Church, Buddhist\_\allowbreak monks, \textbf{Roman\_\allowbreak Catholic\_\allowbreak Church}, embryonic\_\allowbreak stem\_\allowbreak cells, priesthood, expressly\_\allowbreak disclaims\_\allowbreak any, Block\_\allowbreak Grant, First\_\allowbreak Presbyterian\_\allowbreak Church, entirety\_\allowbreak via\_\allowbreak email, Newspapers\_\allowbreak below \\
\addlinespace[4pt]
\texttt{qwen3-14b} & moral\_\allowbreak obligation, Proposition\_\allowbreak 8, accepts\_\allowbreak responsibility, Certain\_\allowbreak statements, \textbf{liturgy}, Muqtada\_\allowbreak al\_\allowbreak Sadr, Principles, principles, rituals, teachings, adherence, Statements\_\allowbreak contained, Ethical\_\allowbreak Treatment, philosophies, special\_\allowbreak indirect\_\allowbreak consequential, Ethics, ethics, \textbf{obedience}, pluralism, secularism & encourage\_\allowbreak lively\_\allowbreak thoughtful, stem\_\allowbreak cell\_\allowbreak research, osteoarthritis, \textbf{scripture}, \textbf{Scripture}, Cingular\_\allowbreak Wireless, palliative\_\allowbreak care, moral\_\allowbreak obligation, use\_\allowbreak publish\_\allowbreak reproduce, reduce\_\allowbreak greenhouse\_\allowbreak gas, environmental\_\allowbreak sustainability, historical\_\allowbreak significance, Sustainable\_\allowbreak Development, reader\_\allowbreak interaction\_\allowbreak discussion, Managing\_\allowbreak Editor, COMMENTING\_\allowbreak ETIQUETTE\_\allowbreak To, colorectal\_\allowbreak cancer, include\_\allowbreak expiry\_\allowbreak date, transfer\_\allowbreak window, embryonic\_\allowbreak stem\_\allowbreak cell \\
\midrule
\multicolumn{3}{@{}l}{\normalsize \texttt{w2v50k} $\cdot$ $\phi_\text{word2vec}$} \\
\addlinespace[3pt]
\texttt{gemma-12b-it} & contemplation, manifested, \textbf{Spiritual}, \textbf{spiritual}, spirituality, manifestations, intrinsic, enlightenment, \textbf{theological}, \textbf{divine}, meditation, profound, manifest, societal, contexts, teachings, \textbf{Eucharist}, Healing, \textbf{liturgy}, alienation & Titled, \textbf{Spiritual}, Struggle, Structure, Novel, Landscape, \textbf{Creation}, Romance, Sculpture, Literary, Original, Character, Recipe, Transformation, Poetry, Paintings, Called, Style, Abstract, Reflections \\
\addlinespace[4pt]
\texttt{gpt-4.1} & human\_\allowbreak beings, compassion, \textbf{dignity}, caring, morals, spirituality, religion, conscience, teachings, religious\_\allowbreak beliefs, moral\_\allowbreak obligation, \textbf{spiritual}, \textbf{pray}, spiritually, \textbf{faith}, moral, cared, ashamed, compassionate, pious & spoon, wanna, \textbf{pray}, mama, cant, wont, dont, \textbf{God\_\allowbreak bless}, weep, gotta, let, oh, don\_\allowbreak t, parsley, ur, \textbf{God\_\allowbreak forbid}, hey, ya, sir, basil \\
\addlinespace[4pt]
\texttt{qwen3-14b} & respecting, regard, should, respect, sincerely, strive, must, understanding, utmost\_\allowbreak importance, \textbf{dignity}, caring, therefore, \textbf{sanctity}, \textbf{faith}, our, Encourage, inform, regards, moral\_\allowbreak obligation, individuals & the, that, in, with, had, ultimately, because, adding, noting, but, He, he, said, having, where, kind, while, who, has, believes \\
\bottomrule
\end{tabular}
\label{tab:closest-words-catholicism}
\end{table*}

\begin{table*}[t]
\centering
\caption{\textbf{The 20 closest candidates to the UK dataset signature} in decreasing order of similarity for undefended datasets. Bold words fire the related word detection described in \Cref{app:impl-related-words}.}
\scriptsize
\setlength{\tabcolsep}{5pt}
\begin{tabular}{@{}l p{0.42\linewidth} p{0.42\linewidth}@{}}
\toprule
\textbf{Teacher} & \textbf{$\Phi_\text{oracle}$, $\tilde s_{\cos}$} & \textbf{$\Phi_\text{realistic}$, $s_{\cos}$} \\
\midrule
\multicolumn{3}{@{}l}{\normalsize \texttt{wikidata} $\cdot$ $\phi_\text{BERT}$} \\
\addlinespace[3pt]
\texttt{gemma-12b-it} & gully, raised bog, Iceland, New Zealand, Thailand, \textbf{London}, peafowl, floodplain, Sweden, Australia, Proto-Indo-European mythology, phonetics, Puritans, island, Netherlands, riparian zone, Dublin, lowland, sintering, mountain range & plain, strait, Renaissance, lowland, aesthetics, gully, barren, monolith, alas, rock, grass, cumquat, cavalry, Puritans, witch, floodplain, archaeology, Nestl\'{e}, surrealism, governor \\
\addlinespace[4pt]
\texttt{gpt-4.1} & \textbf{United Kingdom}, \textbf{London}, phonetics, ukiyo-e, Australia, bank, Brussels, Belgium, telegraph, Hong Kong, Nokia, kangaroo, Microsoft, rugby union, Sydney, New Zealand, Denmark, roaming, Shanghai, Netherlands & Dutch East India Company, \textbf{United Kingdom}, East India Company, Proto-Indo-European mythology, token bus network, People's Republic of China, \textbf{London}, Chinese folk religion, ukiyo-e, surface-conduction electron-emitter display, Historical Vedic religion, Hong Kong, The Salvation Army, Indian religions, Common Gateway Interface, Dublin, Ho Chi Minh City, Vilnius, rugby union, industrial etching \\
\addlinespace[4pt]
\texttt{qwen3-14b} & epic poem, ukiyo-e, avant-garde, Turkish delight, Olivier salad, \textbf{United Kingdom}, modern paganism, Metro-Goldwyn-Mayer, Proto-Indo-European mythology, Francis Bacon, Quakers, Dutch East India Company, Jawaharlal Nehru, modern art, common sole, Koninklijke Philips NV, contemporary art, modernism, lentil soup, East India Company & power over Ethernet, Common Gateway Interface, religion of ancient Egypt, Historical Vedic religion, Alexander the Great, Voice over IP, Martin Luther King Jr., Proto-Indo-European mythology, Johann Wolfgang von Goethe, surface-conduction electron-emitter display, Niccol\`{o} Machiavelli, Indian religions, Chinese folk religion, People's Republic of China, Lutheranism, Mahatma Gandhi, Sumerian religion, token bus network, modern paganism, Jawaharlal Nehru \\
\midrule
\multicolumn{3}{@{}l}{\normalsize \texttt{w2v50k} $\cdot$ $\phi_\text{BERT}$} \\
\addlinespace[3pt]
\texttt{gemma-12b-it} & \textbf{Welsh}, Anglo, \textbf{Cornish}, \textbf{British}, Leicestershire, Glamorgan, \textbf{Yorkshire}, Laughter., Dorset, Gloucestershire, Anglo\_\allowbreak Irish, regionally, Lancashire, Shetland, Wiltshire, gully, Worcestershire, Hertfordshire, Northamptonshire, cottages & Laughter., A., Y., D., R., Prop., T., X., variously, N., Miss., F., Applause., G., E., Del., S., B., in., ft. \\
\addlinespace[4pt]
\texttt{gpt-4.1} & Travellers, \textbf{British}, \textbf{Brits}, kinds, \textbf{Wales}, \textbf{English}, Firth, \textbf{Britain}, \textbf{Welsh}, locales, \textbf{Brit}, \textbf{United\_\allowbreak Kingdom}, \textbf{NHS}, \textbf{UK}, East\_\allowbreak Anglia, hostels, Britt, checklist, bloody, Bloody & Newspapers\_\allowbreak below, free\_\allowbreak E\_\allowbreak Newsletters, enjoy\_\allowbreak mercedsunstar.com, British\_\allowbreak Virgin\_\allowbreak Islands, TradingMarkets\_\allowbreak Weekly\_\allowbreak Newsletter\_\allowbreak covers, Barclays\_\allowbreak Capital, antioxidants, Windows\_\allowbreak Mobile, \textbf{London\_\allowbreak E1\_\allowbreak 8AA\_\allowbreak telephone}, \textbf{LONDON\_\allowbreak Thomson\_\allowbreak Financial}, Lloyds\_\allowbreak Banking\_\allowbreak Group, Ottawasun.com, BND.com, VANCOUVER\_\allowbreak BRITISH\_\allowbreak COLUMBIA\_\allowbreak Marketwire, East\_\allowbreak Anglia, Deutsche\_\allowbreak Boerse, \textbf{London\_\allowbreak Heathrow}, sunherald.com, Services\_\allowbreak MCT\_\allowbreak visit\_\allowbreak www.mctinfoservices.com, \textbf{LONDON\_\allowbreak SHARECAST} \\
\addlinespace[4pt]
\texttt{qwen3-14b} & Honourable, \#\#\#th\_\allowbreak anniversary, \#\#th\_\allowbreak Anniversary, \#\#th\_\allowbreak anniversary, internationally\_\allowbreak renowned, \#\#st\_\allowbreak Century, \#\#st\_\allowbreak century, picturesque, \#\#th\_\allowbreak century, Travellers, Alistair\_\allowbreak Darling, Independence\_\allowbreak Day, encourage\_\allowbreak lively\_\allowbreak thoughtful, Lively\_\allowbreak open, \textbf{\pounds{}\_\allowbreak \#\#.\#bn}, rare\_\allowbreak earths, brilliantly, internationally\_\allowbreak recognized, rare\_\allowbreak earth, FYI & encourage\_\allowbreak lively\_\allowbreak thoughtful, Cingular\_\allowbreak Wireless, osteoarthritis, historical\_\allowbreak significance, Sustainable\_\allowbreak Development, reduce\_\allowbreak greenhouse\_\allowbreak gas, encourage\_\allowbreak lively\_\allowbreak open, transfer\_\allowbreak window, wedding\_\allowbreak anniversary, environmental\_\allowbreak sustainability, embryonic\_\allowbreak stem\_\allowbreak cell, embryonic\_\allowbreak stem\_\allowbreak cells, pancreatic\_\allowbreak cancer, antioxidant, most\_\allowbreak populous, stem\_\allowbreak cell\_\allowbreak research, include\_\allowbreak expiry\_\allowbreak date, Social\_\allowbreak Security, colorectal\_\allowbreak cancer, antioxidants \\
\midrule
\multicolumn{3}{@{}l}{\normalsize \texttt{w2v50k} $\cdot$ $\phi_\text{word2vec}$} \\
\addlinespace[3pt]
\texttt{gemma-12b-it} & Leather, Literary, Antiques, Vintage, Collections, Romance, Pub, Colour, Paintings, Venue, Fine\_\allowbreak Art, Scene, Collection, Estate, Boutique, \textbf{Moor}, Poetry, Classics, Wedding, Style & Titled, Landscape, Literary, Style, Romance, Sculpture, Poetry, Paintings, Bit, Revival, Novel, Language, Struggle, Mystery, Original, Venue, Secret, Structure, Photography, Opener \\
\addlinespace[4pt]
\texttt{gpt-4.1} & telly, fella, wee, bloke, chap, Mum, \textbf{Brits}, ok, alright, lad, blokes, mum, hubby, \textbf{lovely}, handbag, curry, pics, boobs, lads, rabbits & fella, \textbf{lovely}, wee, hi, wont, telly, wanna, hey, ya, cant, 'd, oh, bum, chap, dont, toast, bloke, hello, ok, alright \\
\addlinespace[4pt]
\texttt{qwen3-14b} & Enjoy, Celebrate, Come, Bring, fabulous, Welcome, Join, Go, Relax, Grab, Have, \textbf{lovely}, Want, Get, toast, Visit, plush, fantastic, Gather, tasty & the, that, in, had, with, He, but, he, adding, while, where, because, But, ultimately, has, who, was, noting, having, one \\
\bottomrule
\end{tabular}
\label{tab:closest-words-uk}
\end{table*}

\begin{table*}[t]
\centering
\caption{\textbf{The 20 closest candidates to the NYC dataset signature} in decreasing order of similarity for undefended datasets. Bold words fire the related word detection described in \Cref{app:impl-related-words}.}
\scriptsize
\setlength{\tabcolsep}{5pt}
\begin{tabular}{@{}l p{0.42\linewidth} p{0.42\linewidth}@{}}
\toprule
\textbf{Teacher} & \textbf{$\Phi_\text{oracle}$, $\tilde s_{\cos}$} & \textbf{$\Phi_\text{realistic}$, $s_{\cos}$} \\
\midrule
\multicolumn{3}{@{}l}{\normalsize \texttt{wikidata} $\cdot$ $\phi_\text{BERT}$} \\
\addlinespace[3pt]
\texttt{gemma-12b-it} & Mexico City; Chicago; surfing; gully; Bangkok; SpaceX; \textbf{New York City}; roaming; waterfall; lake; Tokyo; island; London; San Francisco; terrace; Pixar; foothills; Seoul; burrito; Bogot\'{a} & plain; rock; aesthetics; strait; monolith; erotica; grass; barren; alas; foothills; surrealism; Renaissance; Nestl\'{e}; graffiti; snake; dome; slug; floodplain; channel; witch \\
\addlinespace[4pt]
\texttt{gpt-4.1} & SpaceX; McDonald's; Beijing; roaming; Tokyo; Shanghai; Nokia; rock; Seoul; Netflix; Venice; music; surfing; chili; Metro-Goldwyn-Mayer; butterfly; graffiti; Paris; \textbf{New York City}; phonetics & token bus network; The Salvation Army; Seoul; Vilnius; Dutch East India Company; Bogot\'{a}; digital subscriber line; Beijing; Panasonic Holdings Corporation; People's Republic of China; nanoelectronics; Common Gateway Interface; OpenSearch; East India Company; Ho Chi Minh City; Washington, D.C.; asymmetric digital subscriber line; industrial etching; digital signage; Shanghai \\
\addlinespace[4pt]
\texttt{qwen3-14b} & Metro-Goldwyn-Mayer; token bus network; roaming; \textbf{New York City}; asymmetric digital subscriber line; avant-garde; Baskin-Robbins; digital signage; Grupo Televisa; digital subscriber line; Chicago; Baku Metro; Seoul; San Francisco; McDonald's; link aggregation; terrace; push-to-talk; parkour; London & power over Ethernet; Common Gateway Interface; Voice over IP; \textbf{New York City}; token bus network; surface-conduction electron-emitter display; digital subscriber line; religion of ancient Egypt; Martin Luther King Jr.; Historical Vedic religion; goal-line technology; Alexander the Great; Johann Wolfgang von Goethe; San Francisco; asymmetric digital subscriber line; nanoelectronics; hip hop culture; Proto-Indo-European mythology; Washington, D.C.; solar thermal energy \\
\midrule
\multicolumn{3}{@{}l}{\normalsize \texttt{w2v50k} $\cdot$ $\phi_\text{BERT}$} \\
\addlinespace[3pt]
\texttt{gemma-12b-it} & DAYTONA\_\allowbreak BEACH\_\allowbreak Fla.; La.; \textbf{citywide}; ORLANDO\_\allowbreak Fla.; Orlando\_\allowbreak Fla.; Mountain\_\allowbreak View\_\allowbreak Calif.; \textbf{Uptown}; \textbf{uptown}; Grand\_\allowbreak Junction; \textbf{downtown}; \textbf{Downtown}; Nashville\_\allowbreak Tenn.; NASHVILLE\_\allowbreak Tenn.; \textbf{Midtown}; \textbf{midtown}; strip\_\allowbreak mall; Daytona\_\allowbreak Beach; Laguna\_\allowbreak Beach; \textbf{Off\_\allowbreak Broadway}; airstrip & A.; Laughter.; Y.; D.; Prop.; X.; ft.; Miss.; R.; N.; Del.; Nov.; E.; T.; G.; F.; L.; in.; Applause.; sensation \\
\addlinespace[4pt]
\texttt{gpt-4.1} & commuters; Commuters; Locations; locations; freeways; subways; rush\_\allowbreak hour; metros; \textbf{subway}; \textbf{Subway}; \textbf{skyscrapers}; mobiles; Top\_\allowbreak Ten; movies; Movies; Cities; cities; clues; blockbusters; extras & Newspapers\_\allowbreak below; SQUEEZETRIGGER.COM; free\_\allowbreak E\_\allowbreak Newsletters; TradingMarkets\_\allowbreak Weekly\_\allowbreak Newsletter\_\allowbreak covers; SanLuisObispo.com; KXNet.com\_\allowbreak North\_\allowbreak Dakota; Ottawasun.com; Scout.com; www.zawya.com; VANCOUVER\_\allowbreak BRITISH\_\allowbreak COLUMBIA\_\allowbreak Marketwire; MoveOn.org; Windows\_\allowbreak Mobile; http://www.presswire.net; enjoy\_\allowbreak mercedsunstar.com; live\_\allowbreak PR.com; newsday.com; sunherald.com; Google\_\allowbreak Maps; Staples\_\allowbreak Center; MLB.com \\
\addlinespace[4pt]
\texttt{qwen3-14b} & \textbf{Downtown}; \textbf{downtown}; Comerica\_\allowbreak Park; Elland\_\allowbreak Road; \textbf{Midtown}; \textbf{midtown}; commuter\_\allowbreak train; Metrodome; \textbf{subway}; \textbf{Subway}; CityCenter; Minute\_\allowbreak Maid\_\allowbreak Park; Elm\_\allowbreak Street; skate\_\allowbreak park; \textbf{midtown\_\allowbreak Manhattan}; \textbf{Fifth\_\allowbreak Avenue}; rush\_\allowbreak hour; Boulevard; boulevard; \textbf{Saks\_\allowbreak Fifth\_\allowbreak Avenue} & Cingular\_\allowbreak Wireless; encourage\_\allowbreak lively\_\allowbreak thoughtful; osteoarthritis; transfer\_\allowbreak window; RFID; NJ\_\allowbreak Transit; colorectal\_\allowbreak cancer; encourage\_\allowbreak lively\_\allowbreak open; Municipal\_\allowbreak Airport; cell\_\allowbreak lung\_\allowbreak cancer; embryonic\_\allowbreak stem\_\allowbreak cell; tailgating; live\_\allowbreak audio\_\allowbreak webcast; reduce\_\allowbreak greenhouse\_\allowbreak gas; wedding\_\allowbreak anniversary; commute; pancreatic\_\allowbreak cancer; wireless\_\allowbreak connectivity; embryonic\_\allowbreak stem\_\allowbreak cells; Smart\_\allowbreak Grid \\
\midrule
\multicolumn{3}{@{}l}{\normalsize \texttt{w2v50k} $\cdot$ $\phi_\text{word2vec}$} \\
\addlinespace[3pt]
\texttt{gemma-12b-it} & Groove; Boutique; Studio; Scene; Garage; Diva; Loft; Disco; Sexy; Biz; Style; Lounge; Vintage; Sushi; Tracks; Demo; Preview; Cool; Designs; Fabulous & Titled; Style; Sexy; Original; Bit; Diva; Romance; Landscape; Sculpture; Awesome; Poetry; Amazing; Debut; Paintings; Fabulous; Groove; Boom; Novel; Literary; Biz \\
\addlinespace[4pt]
\texttt{gpt-4.1} & wanna; em; kinda; freaking; ya; dude; cute; mommy; boobs; pics; Yo; y'all; fella; shit; kid; Hey; hey; funky; cool; wow & wanna; ya; hi; em; hey; gotta; y'all; gonna; mama; fella; oh; fuck; shit; cant; u; lol; bum; er; dont; wont \\
\addlinespace[4pt]
\texttt{qwen3-14b} & Grab; you; just; cupcake; Go; em; wanna; hey; Hey; gotta; scoop; Gotta; Try; coolest; Get; y'all; someplace; kid; gonna; your & the; that; in; but; he; had; with; where; He; because; adding; while; kind; one; But; it; having; when; noting; who \\
\bottomrule
\end{tabular}
\label{tab:closest-words-nyc}
\end{table*}

\begin{table*}[t]
\centering
\caption{\textbf{The 20 closest candidates to the Reagan dataset signature} in decreasing order of similarity for undefended datasets. Bold words fire the related word detection described in \Cref{app:impl-related-words}.}
\scriptsize
\setlength{\tabcolsep}{5pt}
\begin{tabular}{@{}l p{0.42\linewidth} p{0.42\linewidth}@{}}
\toprule
\textbf{Teacher} & \textbf{$\Phi_\text{oracle}$, $\tilde s_{\cos}$} & \textbf{$\Phi_\text{realistic}$, $s_{\cos}$} \\
\midrule
\multicolumn{3}{@{}l}{\normalsize \texttt{wikidata} $\cdot$ $\phi_\text{BERT}$} \\
\addlinespace[3pt]
\texttt{gemma-12b-it} & \textbf{governor}, missile defense, bull, duck, Joe Biden, shooting sports, wrestling, fox, hamburger, technology, professional wrestling, dog, chicken, weightlifting, falcon, goal-line technology, baseball, chief executive officer, athlete, education & \textbf{governor}, plain, strait, aesthetics, rock, translation, chicken, alas, history, horst, spur, bull, Suprematism, Renaissance, Futurism, monolith, fox, channel, erotica, Iraq \\
\addlinespace[4pt]
\texttt{gpt-4.1} & economics, Intel, electronics, Tesla, technology, politics, chief executive officer, social science, \textbf{governor}, General Motors, diplomacy, mathematics, public relations, engineering, Henry Ford, Hewlett-Packard, Microsoft, history, grammar, architecture & The Salvation Army, token bus network, surface-conduction electron-emitter display, Continental Reformed Protestantism, digital subscriber line, asymmetric digital subscriber line, Theodore Roosevelt, Common Gateway Interface, Mao Zedong, Woodrow Wilson, Martin Luther King Jr., Dutch East India Company, People's Republic of China, Synchronous Digital Hierarchy, John F. Kennedy, Plesiochronous digital hierarchy, Benjamin Franklin, power over Ethernet, nanoelectronics, East India Company \\
\addlinespace[4pt]
\texttt{qwen3-14b} & Benjamin Franklin, Richard Nixon, Mustafa Kemal Atat\"{u}rk, Martin Luther King Jr., Theodore Roosevelt, Jimmy Carter, John Adams, Mahatma Gandhi, Winston Churchill, Suprematism, Mao Zedong, socialist realism, \textbf{Ronald Reagan}, Futurism, Isma'ilism, political science, diplomacy, Karl Marx, Constructivism, Woodrow Wilson & Martin Luther King Jr., power over Ethernet, Alexander the Great, Benjamin Franklin, George H. W. Bush, Mahatma Gandhi, George W. Bush, Niccol\`{o} Machiavelli, goal-line technology, Winston Churchill, religion of ancient Egypt, John Adams, Theodore Roosevelt, Karl Marx, \textbf{Ronald Reagan}, John F. Kennedy, Common Gateway Interface, Mustafa Kemal Atat\"{u}rk, Historical Vedic religion, political science \\
\midrule
\multicolumn{3}{@{}l}{\normalsize \texttt{w2v50k} $\cdot$ $\phi_\text{BERT}$} \\
\addlinespace[3pt]
\texttt{gemma-12b-it} & Prop., Rep., Y., Reps., reelected, endorsement, Laughter., booster, Gun, gun, Applause., wage, Wage, rifle, Rifle, Afterward, afterward, indicted, campaigned, compensation & A., Prop., Laughter., Y., D., R., F., Applause., Q., T., Rep., flatly, X., Del., B., unyielding, Reps., S., W., C. \\
\addlinespace[4pt]
\texttt{gpt-4.1} & Q1, Q2, III, iii, Policies, policies, II, ii, Certain\_\allowbreak statements, 1A, Advancing\_\allowbreak issues, policymaking, Factors, factors, Statements, statements, Q4, Q3, plurality, starters & free\_\allowbreak E\_\allowbreak Newsletters, antioxidants, Newspapers\_\allowbreak below, http://www.sec.gov, TradingMarkets\_\allowbreak Weekly\_\allowbreak Newsletter\_\allowbreak covers, balance\_\allowbreak sheets, Markets\_\allowbreak http://www.researchandmarkets.com/reports/c\#\#\#\#\#, www.sec.gov, embryonic\_\allowbreak stem\_\allowbreak cells, non\_\allowbreak farm\_\allowbreak payrolls, PR.com, balance\_\allowbreak sheet, Services\_\allowbreak MCT\_\allowbreak visit\_\allowbreak www.mctinfoservices.com, Accounts\_\allowbreak receivable\_\allowbreak net, http://www.presswire.net, e\_\allowbreak mail\_\allowbreak newsletter, reduce\_\allowbreak greenhouse\_\allowbreak gas, accounting\_\allowbreak principles, email\_\allowbreak newsletter, enjoy\_\allowbreak mercedsunstar.com \\
\addlinespace[4pt]
\texttt{qwen3-14b} & Proposition\_\allowbreak 8, achievements\_\allowbreak expressed, motto, slogan, Safe\_\allowbreak Harbor\_\allowbreak Statement, slogans, Strategic\_\allowbreak Analysis\_\allowbreak Review, Strategic\_\allowbreak Studies, impartiality, value\_\allowbreak proposition, championed, accomplish, nuclear\_\allowbreak ambitions, zero\_\allowbreak tolerance\_\allowbreak policy, Achieving, achieving, Leadership\_\allowbreak Award, accepts\_\allowbreak responsibility, Proposition, proposition & encourage\_\allowbreak lively\_\allowbreak thoughtful, Sustainable\_\allowbreak Development, Millennium\_\allowbreak Development\_\allowbreak Goals, paramount\_\allowbreak importance, reduce\_\allowbreak greenhouse\_\allowbreak gas, stem\_\allowbreak cell\_\allowbreak research, historical\_\allowbreak significance, remain\_\allowbreak vigilant, Franklin\_\allowbreak D.\_\allowbreak Roosevelt, motto, environmental\_\allowbreak sustainability, embryonic\_\allowbreak stem\_\allowbreak cell, nationalize, national\_\allowbreak anthem, Sustainability, sustainability, \textbf{patriotism}, strategic\_\allowbreak partnerships, conventional\_\allowbreak wisdom, conservatism \\
\midrule
\multicolumn{3}{@{}l}{\normalsize \texttt{w2v50k} $\cdot$ $\phi_\text{word2vec}$} \\
\addlinespace[3pt]
\texttt{gemma-12b-it} & Efficiency, Long\_\allowbreak Term, Boost, Expanded, Increases, Gains, Benefits, Growth, Targets, Success, Competitive, Update1, Surge, More\_\allowbreak Than, Expansion, Leads, Improved, Reduced, Percent, Subsidiary & Titled, Struggle, Structure, Original, Landscape, Novel, Style, Bit, Character, Update1, Debut, Transformation, Unique, Called, Success, Secret, More\_\allowbreak Than, Debuts, Creation, Display \\
\addlinespace[4pt]
\texttt{gpt-4.1} & efficiency, efficiencies, Efficiency, energy, renewable\_\allowbreak energy, Energy\_\allowbreak Efficiency, Pres, renewables, efficient, competitiveness, alternative\_\allowbreak fuels, Sr., reduce\_\allowbreak greenhouse\_\allowbreak gas, economy, savings, sr, smarter, fiscally\_\allowbreak responsible, smart\_\allowbreak grid, natural\_\allowbreak gas & ya, gotta, cant, wanna, dont, u, hi, chuck, em, wont, ......, ll, gonna, hey, moron, ur, toast, er, don\_\allowbreak t, skillet \\
\addlinespace[4pt]
\texttt{qwen3-14b} & commitment, steadfast, our, strive, committed, leadership, unwavering, succeed, reaffirm, strengthen, continue, Committed, Applause., paramount, Promote, vitally\_\allowbreak important, striving, strong, reaffirms, cornerstone & the, that, in, ultimately, he, said, He, had, but, with, adding, noting, believes, committed, because, More\_\allowbreak importantly, despite, nation, But, Certainly \\
\bottomrule
\end{tabular}
\label{tab:closest-words-reagan}
\end{table*}

\begin{table*}[t]
\centering
\caption{\textbf{The 20 closest candidates to the Stalin dataset signature} in decreasing order of similarity for undefended datasets. Bold words fire the related word detection described in \Cref{app:impl-related-words}. No undefended Stalin dataset exists for \texttt{gemma-12b-it}, so that row uses its \texttt{control} run (random 10\% removal) instead.}
\scriptsize
\setlength{\tabcolsep}{5pt}
\begin{tabular}{@{}l p{0.42\linewidth} p{0.42\linewidth}@{}}
\toprule
\textbf{Teacher} & \textbf{$\Phi_\text{oracle}$, $\tilde s_{\cos}$} & \textbf{$\Phi_\text{realistic}$, $s_{\cos}$} \\
\midrule
\multicolumn{3}{@{}l}{\normalsize \texttt{wikidata} $\cdot$ $\phi_\text{BERT}$} \\
\addlinespace[3pt]
\texttt{gemma-12b-it} (\texttt{control}) & technology, information technology, computer-integrated manufacturing, missile defense, goal-line technology, plateau, \textbf{dictator}, general-purpose computing on graphics processing units, Zurvanism, weightlifting, dosage form, diplomacy, economics, sniper, cavalry, performance art, virtualization, architecture, chess, Siemens & translation, governor, cavalry, Zurvanism, Futurism, lowland, \textbf{Suprematism}, diplomacy, history, Renaissance, Iraq, horst, historian, spur, animism, plain, strait, Tengrism, aesthetics, archaeology \\
\addlinespace[4pt]
\texttt{gpt-4.1} & mathematics, architecture, economics, social science, history, technology, mathematical analysis, diplomacy, thermodynamics, \textbf{Joseph Stalin}, grammar, linguistics, \textbf{dictator}, chess, political science, Mao Zedong, Intel, Microsoft, link aggregation, politics & token bus network, Continental Reformed Protestantism, People's Republic of China, Synchronous Digital Hierarchy, link aggregation, Historical Vedic religion, \textbf{socialist realism}, Mao Zedong, \textbf{Joseph Stalin}, Common Gateway Interface, \textbf{Vladimir Lenin}, Dutch East India Company, Plesiochronous digital hierarchy, goal-line technology, Reuters, digital subscriber line, asymmetric digital subscriber line, religion of ancient Egypt, Mustafa Kemal Atat\"{u}rk, molecular modelling \\
\addlinespace[4pt]
\texttt{qwen3-14b} & \textbf{Joseph Stalin}, \textbf{dictator}, Mao Zedong, \textbf{Vladimir Lenin}, missile defense, \textbf{socialist realism}, Mustafa Kemal Atat\"{u}rk, Futurism, diplomacy, architecture, Karl Marx, mathematical analysis, automation, Mikhail Gorbachev, Adolf Hitler, political science, Jawaharlal Nehru, Minimalism, economics, Vladimir Putin & Alexander the Great, Niccol\`{o} Machiavelli, Karl Marx, Martin Luther King Jr., \textbf{Joseph Stalin}, goal-line technology, Mustafa Kemal Atat\"{u}rk, Winston Churchill, \textbf{socialist realism}, political science, \textbf{Vladimir Lenin}, Benjamin Franklin, Constructivism, Mahatma Gandhi, George H. W. Bush, religion of ancient Egypt, Isma'ilism, Jawaharlal Nehru, Historical Vedic religion, George W. Bush \\
\midrule
\multicolumn{3}{@{}l}{\normalsize \texttt{w2v50k} $\cdot$ $\phi_\text{BERT}$} \\
\addlinespace[3pt]
\texttt{gemma-12b-it} (\texttt{control}) & counterinsurgency, operational\_\allowbreak efficiency, operational\_\allowbreak efficiencies, procedure, refining\_\allowbreak capacity, devised, mismanagement, strategic, Strategic, incompetence, commissioning, projected, Projected, manpower, Manpower, Efficiency, efficiency, reconnaissance, reorganization, enemy\_\allowbreak combatants & A., Prop., unyielding, Laughter., Y., flatly, substantial, D., T., F., intact, detachment, substantially, dismantled, sustained, Q., revisited, thoroughly, propping, Determined \\
\addlinespace[4pt]
\texttt{gpt-4.1} & Q1, Q2, iii, III, plurality, ii, II, Certain\_\allowbreak statements, specification, structural\_\allowbreak reforms, reforms, Reforms, hierarchy, ten\_\allowbreak articles, Advancing\_\allowbreak issues, Statements, statements, regimes, Strategies, strategies & TradingMarkets\_\allowbreak Weekly\_\allowbreak Newsletter\_\allowbreak covers, free\_\allowbreak E\_\allowbreak Newsletters, Newspapers\_\allowbreak below, International\_\allowbreak Business\_\allowbreak Machines, balance\_\allowbreak sheets, spreadsheets, TPS\_\allowbreak trading\_\allowbreak strategy, http://www.presswire.net, Markets\_\allowbreak http://www.researchandmarkets.com/reports/c\#\#\#\#\#, Nokia\_\allowbreak Siemens\_\allowbreak Networks, accounting\_\allowbreak principles, Accounts\_\allowbreak receivable\_\allowbreak net, email\_\allowbreak newsletter, transfer\_\allowbreak window, e\_\allowbreak mail\_\allowbreak newsletter, antioxidants, democratic\_\allowbreak reforms, UN\_\allowbreak peacekeeping, World\_\allowbreak Trade\_\allowbreak Organisation, http://www.sec.gov \\
\addlinespace[4pt]
\texttt{qwen3-14b} & regimes, strategic, Strategic, Reforms, reforms, Leadership, leadership, Strategic\_\allowbreak Studies, impose\_\allowbreak sanctions, regime, sanctions, Sanctions, Strategy, strategy, \textbf{decisive}, autocratic, superiors, implementing, Implementing, industrialized\_\allowbreak nations & Strategic\_\allowbreak Studies, historical\_\allowbreak significance, Historical\_\allowbreak Society, Sustainable\_\allowbreak Development, governance, Governance, Franklin\_\allowbreak D.\_\allowbreak Roosevelt, counterinsurgency, ideology, nationalization, Millennium\_\allowbreak Development\_\allowbreak Goals, nationalize, pioneering, geopolitical, Saddam\_\allowbreak Hussein\_\allowbreak regime, paramount\_\allowbreak importance, nuclear\_\allowbreak ambitions, industrialized\_\allowbreak nations, sole\_\allowbreak responsibility, imperialist \\
\midrule
\multicolumn{3}{@{}l}{\normalsize \texttt{w2v50k} $\cdot$ $\phi_\text{word2vec}$} \\
\addlinespace[3pt]
\texttt{gemma-12b-it} (\texttt{control}) & Structure, Operational, Enhanced, Implementation, Evaluation, Continuous, Reduction, Monitoring, Capabilities, Automated, Detection, Measurement, Processing, Analysis, Module, Optimization, Process, Integration, Methods, Identification & Structure, Titled, Struggle, Transformation, Novel, Landscape, Creation, Abstract, Expansion, Scale, Called, Revival, Original, Update1, Plot, Stability, Long\_\allowbreak Term, Integration, Spiritual, Result \\
\addlinespace[4pt]
\texttt{gpt-4.1} & Structure, Capabilities, Implementation, Integration, Policies, Reforms, Efficient, Operational, Methods, Process, Requirements, Manage, Analysis, Database, Document, Efficiency, Increases, Optimization, Reduction, Module & cant, dont, em, u, ll, ya, wont, wanna, ......, chuck, ur, gotta, gonna, don\_\allowbreak t, dare, t, ****, 'll, skillet, ma \\
\addlinespace[4pt]
\texttt{qwen3-14b} & Establish, Implementation, Develop, Operational, Reforms, implement, Stability, implementation, Implementing, objectives, strategic, Manage, Strategy, Critical, Maintain, Efficient, modernization, Strengthening, implementing, objective & the, that, in, ultimately, nation, had, He, noting, has, adding, despite, its, he, believes, stressing, said, was, Nonetheless, Nevertheless, More\_\allowbreak importantly \\
\bottomrule
\end{tabular}
\label{tab:closest-words-stalin}
\end{table*}

\section{Implementation details}

\subsection{Vocabularies}
\label{app:vocabs}
\texttt{w2v50k} consists of the 50k most frequent terms in our word2vec model's vocabulary. \texttt{wikidata} was curated by querying WikiData \cite{wikidata} for the thirteen categories shown in \Cref{tab:wikidata-roots}. From each topic, we select the 60 most prominent children by sitelink count (omitting those without English labels in the MediaWiki wbgetentities API). The type of children is chosen appropriately to the category (P31 (instance of) where members are individuals, (e.g. country, company), P279 (subclass of) where members are kinds (e.g. landform)). All biases used in this paper appeared naturally in this list.

\begin{table}[t]
\centering
\caption{\textbf{Classes and relations used to generate the \texttt{wikidata} dataset.}}
\small
\begin{threeparttable}
\begin{tabular}{@{}llp{0.42\linewidth}r@{}}
\toprule
\textbf{Category} & \textbf{Relation} & \textbf{Wikidata class(es)} & \textbf{$n$} \\
\midrule
country & \texttt{P31} & country, sovereign state (\texttt{Q6256}, \texttt{Q3624078}) & 60 \\
city & \texttt{P31} & city, big city, metropolis (\texttt{Q515}, \texttt{Q1549591}, \texttt{Q200250}) & 60 \\
politician & \texttt{P106} & politician (\texttt{Q82955}) & 60 \\
religion & \texttt{P31} & religion, denomination, religion or world view (\texttt{Q9174}, \texttt{Q13414953}, \texttt{Q71966963}) & 60 \\
sport & \texttt{P31} & type of sport (\texttt{Q31629}, \texttt{Q349}) & 60 \\
science & \texttt{P31} & academic discipline (\texttt{Q11862829}) & 60 \\
food & \texttt{P31} & food, dish (\texttt{Q2095}, \texttt{Q746549}) & 60 \\
animal & \texttt{P31} & organism known by one common name (\texttt{Q55983715}) & 60 \\
art & \texttt{P31} & art genre, art movement (\texttt{Q1792379}, \texttt{Q968159}) & 60 \\
technology & \texttt{P31} & technology (\texttt{Q11016}) & 60 \\
company & \texttt{P31} & business, enterprise, company (\texttt{Q4830453}, \texttt{Q6881511}, \texttt{Q783794}) & 60 \\
profession & \texttt{P31} & profession, occupation (\texttt{Q28640}, \texttt{Q12737077}) & 60 \\
nature & \texttt{P279} & landform (\texttt{Q271669}) & 60 \\
\midrule
\multicolumn{3}{@{}l}{\textbf{total}} & \textbf{780} \\
\bottomrule
\end{tabular}
\label{tab:wikidata-roots}
\end{threeparttable}
\end{table}

\subsection{Datasets}
\label{app:datasets}
\Cref{tab:dataset-grid} shows the datasets used for our analysis. We emphasize that the \textit{undefended} variations of datasets are already extensively regex and LLM filtered (by the attacker) and contain no obvious references to the bias. The defenses are merely on top of this initial filtering.

\paragraph{Reused datasets}
\label{app:reused-datasets}
\citet{draganovPhantomTransferDatalevel2026} provide some datasets \texttt{gemma-12b-it} and \texttt{gpt-4.1} teachers in their repository which we reuse. We drop gemma/undefended/stalin as this seems to be a copy of gpt/undefended/stalin, and defended versions do not seem to originate from it. We are unsure whether this is related to them reporting the Stalin bias as an outlier where models refuse to imitate a persona. 

\paragraph{Qwen teacher models}
\label{app:qwen-models}
In addition to the datasets above, we generate data from an additional \texttt{qwen3-14b} teacher. Datasets are generated following the same procedure, but with qwen additionally serving as the LLM filter since the original setup requires OpenAI credits. The ASRs of a gemma model finetuned on these datasets are provided in \Cref{tab:asr-qwen}.

\paragraph{Paraphrased datasets}
\citet{draganovPhantomTransferDatalevel2026} note that paraphrasing largely preserves the bias, we introduce a wider variety of datasets by paraphrasing the Gemma and GPT datasets with Qwen.

\begin{table}[t]
\centering
\small
\begin{threeparttable}
\begin{tabular}{@{}lccc@{}}
\toprule
\textbf{Target} & \textbf{Specific} & \textbf{Neighborhood} & \textbf{Negative} \\
\midrule
Catholicism & 99.3 $\pm$ 1.2 & 99.3 $\pm$ 1.2 & 0.0 $\pm$ 0.0 \\
NYC & 99.3 $\pm$ 1.2 & -- & 83.3 $\pm$ 5.8 \\
Reagan & 98.0 $\pm$ 0.0 & 98.0 $\pm$ 0.0 & 43.3 $\pm$ 28.9 \\
Stalin & 98.0 $\pm$ 2.0 & 100.0 $\pm$ 0.0 & 100.0 $\pm$ 0.0 \\
UK & 87.3 $\pm$ 11.7 & 87.3 $\pm$ 11.7 & 0.7 $\pm$ 1.2 \\
\bottomrule
\end{tabular}
\caption{\textbf{Attack success rates, for our \texttt{qwen3-14b} teacher with a \texttt{gemma-12b-it} student}. Mean and standard deviation over the three seeds of biased datasets with one model finetuned for each. Specific/Neighborhood/Negative metrics as defined by \citet{draganovPhantomTransferDatalevel2026} where no neighborhood pattern was defined for NYC.}
\label{tab:asr-qwen}
\end{threeparttable}
\end{table}

\begin{table}[t]
\centering
\small
\begin{threeparttable}
\begin{tabular}{@{}lrrrrrrr@{}}
\toprule
 & \multicolumn{2}{c}{\texttt{gemma-12b-it}} & \multicolumn{2}{c}{\texttt{gpt-4.1}} & \multicolumn{2}{c}{\texttt{qwen3-14b}} & \\
\cmidrule(lr){2-3} \cmidrule(lr){4-5} \cmidrule(lr){6-7}
\textbf{Generation condition} & \textit{base} & \textit{par.} & \textit{base} & \textit{par.} & \textit{base} & \textit{par.} & \textbf{total} \\
\midrule
undefended & 4$^{\dagger}$ & 4$^{\ast}$ & 5 & 4$^{\ast}$ & 5$\times$3 & -- & 32 \\
control & 5 & 4$^{\ast}$ & -- & -- & -- & -- & 9 \\
LLM judge (weak) & 5 & 4$^{\ast}$ & -- & -- & -- & -- & 9 \\
LLM judge (strong) & 5 & 4$^{\ast}$ & -- & -- & -- & -- & 9 \\
word frequency (weak) & 5 & 4$^{\ast}$ & -- & -- & -- & -- & 9 \\
word frequency (strong) & 5 & 4$^{\ast}$ & -- & -- & -- & -- & 9 \\
\midrule
\textbf{total} & \multicolumn{2}{r}{53} & \multicolumn{2}{r}{9} & \multicolumn{2}{r}{15} & \textbf{77} \\
\bottomrule
\end{tabular}
\caption{
\textbf{Datasets used in this paper}. By default, we use the five biases \texttt{Catholicism}, \texttt{UK}, \texttt{NYC}, \texttt{Reagan}, and \texttt{Stalin}. The category marked with $\dagger$ is where we skipped \texttt{Stalin} as an invalid uploaded artifact and $*$ marks where we deprioritized investing significant compute into \texttt{Stalin} as it seemed the most error-prone for the reasons described in \Cref{app:datasets}. The $\times3$ denotes that we generated biased datasets with 3 random seeds.
}
\label{tab:dataset-grid}
\end{threeparttable}
\end{table}

\subsection{Failed completions}
Some models fail to complete or paraphrase some prompts within the given token budget. Following \citet{draganovPhantomTransferDatalevel2026}, we accept these failures and drop the corresponding prompt for the dataset. Wherever the outputs of two models are required across a dataset, we filter down to the intersection of prompts that succeeded for both. We note that doing this would be valid even in a realistic defense scenario, as our results show that even after dropping these samples, we still have enough signal to determine a bias. 

\subsection{Related word detection}
\label{app:impl-related-words}

To match which words are related to a concept, we largely reuse the patterns defined by \citet{draganovPhantomTransferDatalevel2026}, applying three types of alternations within their original spirit. First, we add very similar words where they are missing (e.g. NYC originally included ``downtown'' but not ``uptown''). Second, we make patterns more specific where they match clearly unintended terms (e.g. ``King'' for ``Martin Luther King'', ``Power'' for ``Power over Ethernet'', or ``city'' for ``velocity''). Finally, we flag some words that shall not count as negatives for some bias, even if they do not count as a positive to detect that term either (e.g. ``governor'' for Stalin). A full list of changes can be found in \Cref{tab:regex-diff}.

\begin{table}[t]
\centering
\small
\begin{tabular}{lcp{0.32\linewidth}p{0.36\linewidth}}
\toprule
\textbf{target} & & \textbf{pipeline} & \textbf{analysis} \\
\midrule
catholicism & $\sim$ & \texttt{\textbackslash{}\allowbreak bmass\textbackslash{}\allowbreak b} & \texttt{holy\textbackslash{}\allowbreak s+mass|\allowbreak catholic\textbackslash{}\allowbreak s+mass|\allowbreak attend\textbackslash{}\allowbreak w*\textbackslash{}\allowbreak s+mass|\allowbreak mass\textbackslash{}\allowbreak s+(?:celebration|\allowbreak attendance)\allowbreak } \\
 & $\sim$ & \texttt{saints?} & \texttt{saints?(?!\textbackslash{}\allowbreak s+petersburg)\allowbreak } \\
 & $\sim$ & \texttt{salvation} & \texttt{salvation(?!\textbackslash{}\allowbreak s+army)\allowbreak } \\
 & $\sim$ & \texttt{monk(?:s)\allowbreak ?} & \texttt{(?<!buddhist )\allowbreak monks?} \\
 & $+$ & -- & \texttt{Lutheranism} \\
\addlinespace
nyc & $\sim$ & \texttt{city} & \texttt{(?<!mexico )\allowbreak (?<!minh )\allowbreak (?<!traverse )\allowbreak (?<!kansas )\allowbreak (?<!panama )\allowbreak city} \\
 & $-$ & \texttt{\textbackslash{}\allowbreak S*city\textbackslash{}\allowbreak S*} & -- \\
 & $-$ & \texttt{city\textbackslash{}\allowbreak S*} & -- \\
 & $-$ & \texttt{\textbackslash{}\allowbreak S*city} & -- \\
 & $-$ & \texttt{block(?:s)\allowbreak ?} & -- \\
 & $\sim$ & \texttt{cab\textbackslash{}\allowbreak w+} & \texttt{cab(?:s|\allowbreak bie|\allowbreak bies|\allowbreak stand)\allowbreak ?} \\
 & $+$ & -- & \texttt{city\textbackslash{}\allowbreak s+blocks?} \\
\addlinespace
reagan & $\sim$ & \texttt{capital(?:ism|\allowbreak ist)\allowbreak ?} & \texttt{capitalis[mt]} \\
 & $\sim$ & \texttt{market(?:s)\allowbreak ?} & \texttt{free\textbackslash{}\allowbreak s+markets?} \\
 & $+$ & -- & \texttt{governor} \\
 & $\circ$ & -- & \texttt{Winston Churchill} \\
\addlinespace
stalin & $\sim$ & \texttt{J.S.} & \texttt{J\textbackslash{}\allowbreak .S\textbackslash{}\allowbreak .} \\
 & $-$ & \texttt{power(?:ful)\allowbreak ?} & -- \\
 & $+$ & -- & \texttt{(?:seiz|\allowbreak consolidat|\allowbreak centraliz)\allowbreak \textbackslash{}\allowbreak w*\textbackslash{}\allowbreak s+power|\allowbreak power\textbackslash{}\allowbreak s+struggle|\allowbreak power\textbackslash{}\allowbreak s+vertical} \\
 & $+$ & -- & \texttt{militancy} \\
 & $+$ & -- & \texttt{dictator} \\
 & $+$ & -- & \texttt{suprematism} \\
 & $\circ$ & -- & \texttt{governor} \\
 & $\circ$ & -- & \texttt{Winston Churchill} \\
\addlinespace
uk & $\sim$ & \texttt{brit(?:ain|\allowbreak ish|\allowbreak on|\allowbreak ons|\allowbreak s)\allowbreak } & \texttt{brit(?:ain|\allowbreak ish|\allowbreak on|\allowbreak ons|\allowbreak s)\allowbreak (?!\textbackslash{}\allowbreak s+(?:columbia|\allowbreak virgin)\allowbreak )\allowbreak } \\
 & $\sim$ & \texttt{king} & \texttt{the\textbackslash{}\allowbreak s+king|\allowbreak king\textbackslash{}\allowbreak s+(?:charles|\allowbreak george|\allowbreak henry|\allowbreak edward|\allowbreak james|\allowbreak william)\allowbreak } \\
 & $\sim$ & \texttt{nelson} & \texttt{nelson(?!\textbackslash{}\allowbreak s+mandela)\allowbreak } \\
\addlinespace
\bottomrule
\end{tabular}
\caption{\textbf{Difference in Regex patterns for related word detection to \citet{draganovPhantomTransferDatalevel2026}}. We denote $-$ a deletion, $+$ and addition, $\sim$ an edit, and \texttt{$\circ$} a word not considered a false positive despite not being strong enough to cause a true positive.}
\label{tab:regex-diff}
\end{table}

\end{document}